\documentclass[10pt,twocolumn,letterpaper]{article}

\usepackage{cvpr}      

\newcommand{\ApproachName}{{SAP3D}}
\usepackage{url}

\usepackage[accsupp]{axessibility}
\usepackage{pifont}
\usepackage{booktabs}
\usepackage{graphicx}
\usepackage{multirow}
\usepackage{colortbl}
\usepackage{microtype}
\usepackage{gensymb}
\usepackage{pifont}  
\usepackage{multirow, array}
\usepackage{booktabs}  
\usepackage{amsmath, amssymb, bm}

\usepackage[pagebackref,breaklinks,colorlinks,citecolor=cvprblue]{hyperref}
\usepackage{bm}

\newcommand\blfootnote[1]{%
  \begingroup
  \renewcommand\thefootnote{}\footnote{#1}%
  \addtocounter{footnote}{-1}%
  \endgroup
}
\newcommand{\cmark}{\ding{51}}

\def\loss{\mathcal{L}}
\def\paperID{5434} 
\def\confName{CVPR}
\def\confYear{2024}

\title{The More You See in 2D, the More You Perceive in 3D}
\vspace{-1em}

\author{Xinyang Han*\\
UC Berkeley\\
{\tt\small hanxinyang66@gmail.com}
\and
Zelin Gao*\\
Zhejiang University\\
{\tt\small jamesgzl@zju.edu.cn}
\and
Angjoo Kanazawa\\
UC Berkeley\\
{\tt\small kanazawa@berkeley.edu}
\and
Shubham Goel$^{\dagger}$\\
Avataar\\
{\tt\small shubhamgoel@avataar.ai}
\and
Yossi Gandelsman$^{\dagger}$\\
UC Berkeley\\
{\tt\small yossi\_gandelsman@berkeley.com}
}

\begin{document}
\vspace{-1.5em}
\definecolor{cvprblue}{rgb}{0.21,0.49,0.74}
\maketitle
\blfootnote{* Equal contribution. $^\dagger$ Equal contribution.}

\vspace{-1.5em}
\begin{abstract}
\vspace{-0.5em}
Humans can infer 3D structure from 2D images of an object based on past experience and improve their 3D understanding as they see more images. Inspired by this behavior, we introduce SAP3D, a system for 3D reconstruction and novel view synthesis from an arbitrary number of unposed images. Given a few unposed images of an object, we adapt a pre-trained view-conditioned diffusion model together with the camera poses of the images via test-time fine-tuning. The adapted diffusion model and the obtained camera poses are then utilized as instance-specific priors for 3D reconstruction and novel view synthesis. We show that as the number of input images increases, the performance of our approach improves, bridging the gap between optimization-based prior-less 3D reconstruction methods and single-image-to-3D diffusion-based methods. We demonstrate our system on real images as well as standard synthetic benchmarks. Our ablation studies confirm that this adaption behavior is key for more accurate 3D understanding.\footnote{Project page: \url{https://sap3d.github.io/}}
\end{abstract}


\begin{figure}[t!]
\vspace{-0.5em}
    \centering
    \includegraphics[width=0.99\linewidth]{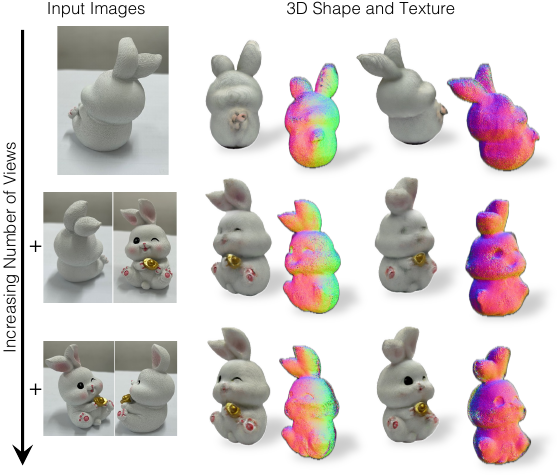}
    \caption{\small \textbf{3D from one or more unposed views.} Our system reconstructs the 3D shape and texture of an object with a variable number of real input images. The first, second, and third rows show reconstructions from 1, 3, and 5 input images. The quality of 3D shape and texture improves with more views.}
    \label{fig:teaser}
\end{figure}
\vspace{-0.5em}
\section{Introduction}
Imagine you are shopping online and see a picture of the bunny in \Cref{fig:teaser} you want to buy. 
You can build a rough mental 3D model of the object based on the first image alone following your understanding of what bunnies look like, 
but you can only guess the geometry and appearance of the unseen parts.
When you see more images, you implicitly estimate the camera viewpoints and consolidate information from all the views to build a better 3D understanding of the object. To accomplish this task you combine 2D priors coming from your vast previous visual experience with the actual observations you see, to understand the 3D object. As you see more examples, your 3D understanding improves.

In this paper, we present a system that follows the same principles - it handles a variable number of images, incorporates learned priors from large-scale 2D image data, and improves as it sees more captures of the object. Specifically, starting from an arbitrary number of input images \textit{without camera poses}, we use a pre-trained generative image model, adapt it to the object of interest, and consolidate the information, enabling consistent 3D reconstruction and novel view synthesis (NVS). Therefore, we call our system {\textbf{S}ee-\textbf{A}dapt-\textbf{P}erceive}, or \ApproachName~in short. 

To reconstruct real-world 3D objects from only a few images (e.g. 3 photos), our system relies on priors from models that are pre-trained on large-scale datasets. We incorporate a generative model that was trained on a large-scale image dataset (LAION~\cite{laion}) and distill it into a 3D model, akin to Zero-1-to-3~\cite{liu2023zero}. We also train a camera pose estimator~\cite{lin2023relpose++}  on a large 3D object dataset (Objaverse~\cite{objaverse}), to enable its generalization to a diverse set of objects. 
The rough camera poses predicted by these models provide us with enough signal for exploiting the generative model prior for 3D reconstruction.
We believe that future improvements in any of the pre-trained models will directly improve our system. 

While other methods that rely on 2D priors use a fixed number of input images, \ApproachName~can incorporate information from a varying number of views. 
To achieve this, we adapt the 2D generative model at test-time by fine-tuning it on the input views and simultaneously refining their estimated relative camera poses. The resulting instance-specific generative model can be used for sampling novel views of the instance and for distillation into a 3D model (e.g. via NeRF~\cite{nerf}).  Thus, our system bridges the gap between single-view and multi-view reconstruction. It can recover 3D geometry and appearance from an \textit{arbitrary number} of un-posed images, as shown in \Cref{fig:teaser}.

As the number of input images grows, the quality of the 3D reconstruction and the estimated camera poses improve. Similarly, the consistency of the novel views sampled from the obtained instance-specific generative model improves with more views. We illustrate this qualitatively for real images and quantitatively for 3D models from the GSO dataset~\cite{gso}. 

Finally, we present an extensive ablation study of different components of our system. We show that test-time adaption, as well as training the camera pose estimator on a large-scale 3D dataset, improves 3D reconstruction, camera pose predictions, and novel view synthesis quality.






\begin{table}[t]
\centering  
\footnotesize  
\begin{tabular}{l@{\hskip4pt}c@{\hskip4pt}@{\hskip4pt}c@{\hskip4pt}@{\hskip4pt}c@{\hskip4pt}@{\hskip4pt}c@{\hskip4pt}@{\hskip4pt}c@{\hskip4pt}}
\toprule  
\begin{tabular}[c]{@{}c@{}}Methods\end{tabular} & 
\begin{tabular}[c]{@{}c@{}} \# Input \\ Views\end{tabular}& 
\begin{tabular}[c]{@{}c@{}}Handles \\ Unposed \\ Images\end{tabular} & 
\begin{tabular}[c]{@{}c@{}}3D \\ Recon.\end{tabular} & 
\begin{tabular}[c]{@{}c@{}}Leverages \\ 2D Data\end{tabular} \\
\midrule  
MV-Dream \cite{shi2023mvdream}           & 0 (text)  &        &         & \cmark \\ 
Zero123 \cite{liu2023zero}             & 1         &        & \cmark  & \cmark \\ 
SyncDreamer \cite{liu2023syncdreamer}          & 1         &        & \cmark  & \cmark \\ 
PixelNeRF \cite{pixelnerf}           & $\geq 1$  &        & \cmark  &        \\ 
SparseFusion \cite{zhou2023sparsefusion}             & $\geq 1$         &        & \cmark  & \cmark \\ 
RelPose++ \cite{lin2023relpose++}           & $\geq 1$  & \cmark & \cmark* &        \\ 
Ours (\ApproachName) & $\geq 1$  & \cmark & \cmark  & \cmark \\ 
\bottomrule  
\end{tabular}
\caption{\small \textbf{Related approaches:} A comparison of different methods, highlighting the key difference in inputs needed, outputs recovered, and type of data used. Only our system (\ApproachName) can reconstruct 3D from a variable number of unposed input images. $^*$Camera outputs from RelPose++ have been shown useful for 3D reconstruction, but only on a limited number of examples with 7 or more images.
}
\label{tab:comparison}
\end{table}

\section{Related Work}

\paragraph{Instance-specific 3D Reconstruction.}
There is extensive literature on 3D reconstructing objects and scenes from input images. COLMAP~\cite{colmap} is the culmination of a long line of classical work on SfM~\cite{pollefeys2004visual,hartley_zisserman_2004} and MVS~\cite{furukawa2009accurate,furukawa2015multi}. NeRF \cite{mildenhall2020nerf}, VolSDF \cite{volsdf}, SRN~\cite{sitzmann2019srns} are modern neural counterparts. However, these approaches do not rely on priors that can be learned from large-scale datasets. Therefore, they require many input views, where camera poses are either known or can be found using SfM. Some approaches \cite{goel2022differentiable, zhang2021ners, Niemeyer2021Regnerf, dietNerf, somraj2022decompnet} attempt to reconstruct from a limited number of views but they need camera poses and cannot reconstruct unseen parts because of the lack of data-driven priors.

\paragraph{Single-view 3D Reconstruction.}
Several works learn priors that enable full 3D reconstruction, including unseen areas. They are often formulated as single-view reconstruction of volumetric occupancy~\cite{choy20163d,occnet,genre,pifuSHNMKL19},  meshes~\cite{groueix2018atlasnet,wang2018pixel2mesh,gkioxari2019mesh} and category-specific shape deformation models~ \cite{kar2015category,kanazawa2018learning,goel2020shape,li2020self,duggal2022tars3D,wu2023dove}, all from a single image.

An alternative two-stage approach for single-view 3D reconstruction relies on implicit 3D priors that 2D generative models learn.   
First, a 2D diffusion model \cite{ramesh2021zero, saharia2022photorealistic, rombach2022high} is pre-trained on a large-scale dataset of images. Second, the pre-trained diffusion model is used as a score function to supervise the optimization of an instance-specific 3D model (e.g. NeRF~\cite{nerf}) via a score distillation sampling (SDS) loss~\cite{poole2022dreamfusion} or a similar variant~\cite{wang2023score}. Zero-1-to-3~\cite{liu2023zero} trains a 2D diffusion model conditioned on a relative camera viewpoint and an input image, and distills the diffusion model into a NeRF via SDS loss by conditioning the diffusion model on a single input image. However, the sampled generated novel views are not multiview-consistent.  Other methods \cite{liu2023syncdreamer, shi2023mvdream} train a diffusion model to generate multiview-consistent images from a single-view image/text, before the distillation. As shown in \Cref{tab:comparison}, all these approaches utilize a single input at test-time, and can not improve the 3D reconstruction by utilizing multiple unposed images of the captured object.  

\begin{figure*}[t]
    \centering
    \includegraphics[width=0.95\linewidth]{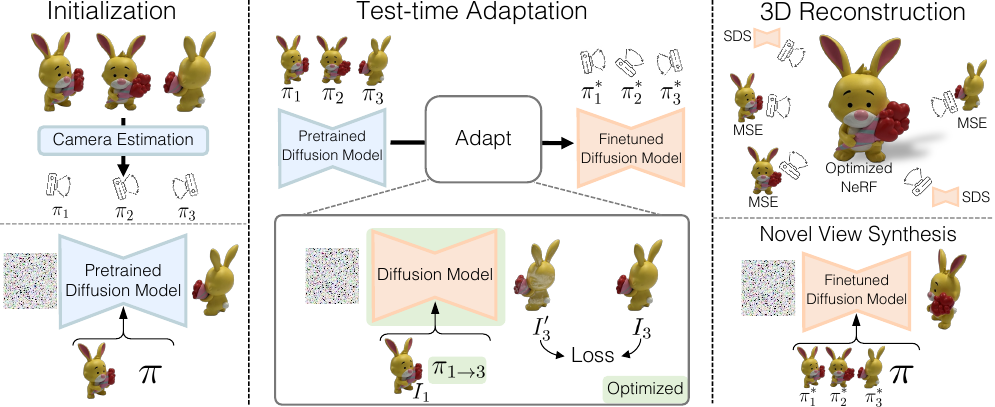}
    \caption{\textbf{Overview of \ApproachName.} We first compute coarse relative camera poses using an off-the-shelf model. We fine-tune a view-conditioned 2D diffusion model on the input images and simultaneously refine the camera poses via optimization. The resulting instance-specific diffusion model and camera poses enable 3D reconstruction and novel view synthesis from \textit{an arbitrary number of input images}. }
    \label{fig:approach}
\end{figure*}

\paragraph{Few-view 3D Reconstruction}
Single-view and multi-view reconstruction are only two ends of the input spectrum. We would like systems that can reconstruct an approximate shape from a single input image but progressively improve with more views. LSM~\cite{LearntStereoMachines} was one of the first to show such behavior in a learning framework, by merging features from a variable number of input images to predict a 3D voxel grid. Recent works~\cite{pixelnerf, co3d} also achieved this with an implicit volumetric representation. However, these approaches require accurate camera poses, and even then they are susceptible to generating blurry outputs due to their deterministic nature.

A recent line of work~\cite{zhang2022relpose,lin2023relpose++,sinha2023sparsepose,wang2023posediffusion} learns to predict relative camera poses from a few input images with small overlap. Even though they are motivated by the goal of sparse view 3D reconstruction, they only show minimal proof-of-concept experiments for 3D reconstruction using their output poses.
Our system, \ApproachName, reconstructs 3D shape from only a few images without the need for input camera poses.

\paragraph{Test-Time Adaptation.} 
 Test-time adaptation aims to fine-tune a pre-trained model on test example(s) before making a prediction. 
 In the context of generative models, and in our case, test-time adaptation is often used for personalizing the model and improving the generation quality on the test distribution. Pivotal Tuning~\cite{DBLP:journals/corr/abs-2106-05744} and MyStyle~\cite{nitzan2022mystyle} allow for real image editing by adapting a generative adversarial network (GAN) to reconstruct face image(s) of a single person. DreamBooth~\cite{dreambooth} and CustomDiffusion~\cite{kumari2022customdiffusion} obtain a personalized text-conditioned model that can synthesize novel images of a subject, by fine-tuning a pre-trained text-to-image diffusion model on a few images of this subject. Dreambooth3D~\cite{raj2023dreambooth3d} distills a personalized text-conditioned model into a NeRF, to enable personalized text-to-3D generation. 
 Similarly, we use test-time adaptation to obtain an \textit{instance-specific} diffusion model, by fine-tuning a view-conditioned diffusion model on a few images of it. Differently from all these methods, our goal is 3D reconstruction and novel view synthesis, which requires preserving the 3D structure of the captured instance. 

\section{\ApproachName}
\label{approach}

We start by describing our three-stage system for 3D object reconstruction and novel view synthesis from a few unposed images, as shown in \ref{fig:approach}. \Cref{approach:initialization} presents the initialization stage of relative camera poses between the input images and a view-conditioned 2D diffusion model that provides implicit 3D priors for the 3D reconstruction. \Cref{approach:ttt} describes the refinement stage of the relative camera poses together with the diffusion model to obtain instance-specific 3D priors via test-time optimization. Finally, we show how the refined camera poses and the instance-specific diffusion model are utilized for 3D reconstruction (\Cref{approach:sds}) and for NVS (\Cref{approach:nvs}).

\subsection{Initialization}
\label{approach:initialization}
\paragraph{Initial camera poses.} Given only a small set of $k$ images of an object $S = \{I_1,...,I_k\}$, $ I_i \in \mathbb{R}^{H \times W \times 3}$, we estimate their corresponding camera poses $\pi_i$ following \cite{lin2023relpose++}. We first compute crude estimates of the distribution of relative poses $\pi_{i \rightarrow j}$ between each pair of images $I_i, I_j$ as a distribution on relative camera rotations $R_{i\rightarrow j} \in \mathbb{R}^{3 \times 3}$ and translations $T_{i\rightarrow j} \in \mathbb{R}^3$. We then find a consistent set of poses $\pi_i$ that maximize the likelihood under the predicted distribution. 
We further refine these camera poses in \Cref{approach:ttt}. 

\paragraph{Initial view-conditioned 2D diffusion model.}
\label{approach:intialization:pretraining}
As 3D reconstruction from a small number of images is an ill-posed problem, we rely on 3D priors that are learned from large-scale image datasets. We use a pre-trained view-conditioned diffusion model $F$ that takes an input object image $I$ and a relative camera pose $\pi$ as conditioning signals, and is trained to generate a new image $I'$ of the object under this camera transformation. $F$ is first pre-trained on an internet-scale dataset of text-image pairs, and then adapted via fine-tuning for view-conditioning, as described in Zero-1-to-3~\cite{liu2023zero}. $F$ consists of a camera pose conditioning network $c_\phi$ and a learned denoiser $\epsilon_\theta$.  It is based on stable diffusion~\cite{ldm} and operates in the latent space of a pre-trained VAE with a fixed encoder $\mathcal{E}$ and decoder $\mathcal{D}$. The diffusion process adds noise to the encoded latent $z = \mathcal{E}(I')$ to produce a noisy latent $z_t$ with an increasing noise level over timesteps $t$. The denoiser $\epsilon_\theta$ learns to predict the noise added to the noisy latent $z_t$ given $I$ and encoded camera pose $c_\phi(\pi)$. Formally, the optimization objective is: 
\begin{equation}
\min _{\theta, \phi} \mathbb{E}_{I,z,\pi,t, \epsilon \sim \mathcal{N}(0,1)}\left\|\epsilon-\epsilon_{\theta}\left(z_{t}, t, c_\phi(\pi), I\right)\right\|_{2}^{2}
\end{equation}
where $\epsilon$ is the noise added to $z$ to create $z_t$.

\subsection{Test-time optimization}
\label{approach:ttt}

\begin{figure*}[h!]
\includegraphics[width=0.99\linewidth]{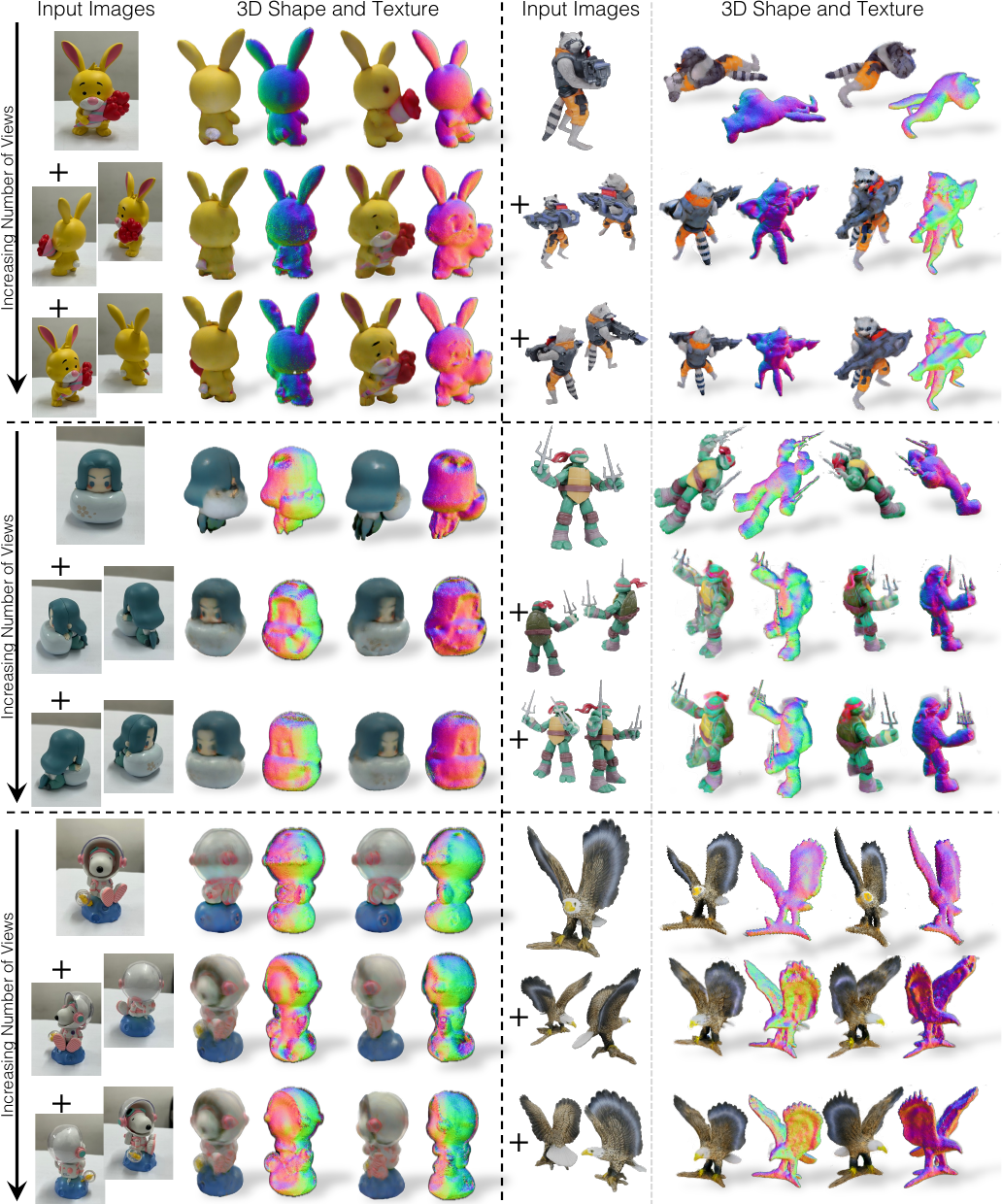}
    \caption{\textbf{3D reconstructions with one or more images.} Qualitative visualizations with 1, 3, and 5 views for \ApproachName~on real images (left column) and instances from the synthetic GSO dataset (right column). Observe how the wings of the eagle, the spiky weapon of the green turtle, and the yellow bunny's bouquet of flowers, all become more detailed and accurate with more views.}
    \label{fig:real}
\end{figure*}

To incorporate information from a varying number of views into our system, we apply test-time optimization on the initial diffusion model and camera poses. Our goals are to (1) obtain an instance-specific view-conditioned diffusion model $F_S$ and (2) improve the camera pose estimates for the images for downstream novel view synthesis and 3D reconstruction. 

\paragraph{Finetuning the diffusion model.} We adapt the diffusion model $F$ to our images $S$ during test-time. We fine-tune $F$ on the images in $S$ using the estimated relative camera poses from between every image pair from the initialization step (resulting in $k(k-1)$ training examples). This stage can be applied with any $k$, as each optimization step requires pairs of images. 


\paragraph{Optimizing camera poses.} During the finetuning process, we also refine the initial estimates for 
$\pi_{i}$
by directly back-propagating to these parameters.
The view-conditioned 2D diffusion model $F$, which can synthesize novel views of an object, models the probability distribution $P(I' | I,\pi)$ of novel views $I'$ of an image $I$ conditioned on the relative camera pose $\pi$. Therefore, by Bayes Rule, it also implicitly models the distribution of camera poses given images $I$ and $I'$. 
We exploit $F$ to further optimize our camera poses. While fine-tuning $F$, we simultaneously optimize our camera pose estimates, by backpropagating gradients into their parameters. 
Formally, the test-time adaptation objective is:
\begin{equation}
\min _{\theta, \phi, \pi_{i}} \mathbb{E}_{i,j,t,\epsilon}\left\|\epsilon-\epsilon_{\theta}\left(z^{j}_{t}, t, c_\phi,(\pi_{i\rightarrow j}), I_i\right)\right\|_{2}^{2}
\end{equation}
where $z^j_t$ is the noisy latent encoding of image $I_j$.

\paragraph{3D prior preservation loss.}
The test-time adaptation of the diffusion model on a few pairs of images can lead to catastrophic forgetting of the learned 3D priors. To address this problem,   
 we incorporate a \textit{3D prior preservation loss} as a regularization term, following DreamBooth~\cite{dreambooth}. We sample pairs of object images and relative camera poses from large-scale object dataset, and include these samples during the test-time optimization. Specifically, we sample images that are similar to the test object from $F$'s training data. We use CLIP~\cite{clip} as a similarity metric and retrieve the nearest neighbors of the average image representation of the images of the object. The loss in test time is:
 \begin{equation}
     \mathcal{L} = \alpha \mathcal{L}_{\text{denoise}} +  \mathcal{L}_{\text{prior}}
 \end{equation}
where $\mathcal{L}_{\text{denoise}}$ is the diffusion loss, $\mathcal{L}_{\text{prior}}$ is the prior preservation loss and $\alpha \in \mathbb{R}$ is the loss coefficient.

\subsection{Novel View Synthesis}
\label{approach:nvs}
The instance-specific diffusion model obtained via test-time optimization allows us to sample novel views of the given object directly. We use stochastic conditioning~\cite{watson2022novel} to incorporate \textit{all of the input images} and the refined camera poses during the sampling process from the diffusion model. The original sampling process uses the same input image and relative camera pose in all of the denoising steps to generate a novel view. Differently, we make use of all the images during the sampling process via stochastic conditioning: 
We first compute the relative camera poses from each of the input images $I_i$ given the refined camera poses $\pi_i$. Then, at each denoising step, we sample a different input image and corresponding camera pose and conditioning $\epsilon_{\theta}$ on it.

\subsection{3D Reconstruction}
\label{approach:sds}


Given the input images $\{I_i\}$ of the object, refined estimated camera poses $\{\pi_i\}$, and the instance-specific diffusion model $F$ that generates novel views of the object, we reconstruct the object in 3D as a neural radiance field \cite{mildenhall2020nerf}  with parameters $\psi$. 
We adapt an existing single-image 3D reconstruction pipeline \cite{liu2023zero} that uses view-conditioned diffusion models, to multiple images.
\begin{table*}[h!]
\begin{minipage}{0.65\linewidth}
\small
    \centering
    \begin{tabular}{c|ccc|cc}
         \#Images&  LPIPS $\downarrow$ &  PSNR $\uparrow$ &  SSIM $\uparrow$ &  CD $\downarrow$&VolumeIoU $\uparrow$\\
         \toprule
         1 (Zero1-to-3)& 0.23 & 14.1 & 0.82 & 0.168 & 0.25 \\
         \midrule
         2 & 0.16 & 18.0 & 0.83 & 0.041 & 0.51\\
         3 & 0.14 & 18.5 & 0.83 & 0.024 & 0.57\\
         4 & 0.11 & 19.6 & 0.85 & 0.023 & 0.67\\
         5 & 0.11 & 19.8 & 0.86 & 0.019 & 0.68\\
         \bottomrule
    \end{tabular}
\vspace{-0.5em}    \caption{\small \textbf{3D reconstruction benchmark.} We compare geometry and appearance accuracy. As more images are provided, the 3D reconstruction quality improves.} 
    \label{tab:more_img_3D}
\end{minipage}\hfill
\begin{minipage}{0.30\linewidth}\vspace{-1em}
\centering
		\includegraphics[width=0.95\textwidth]{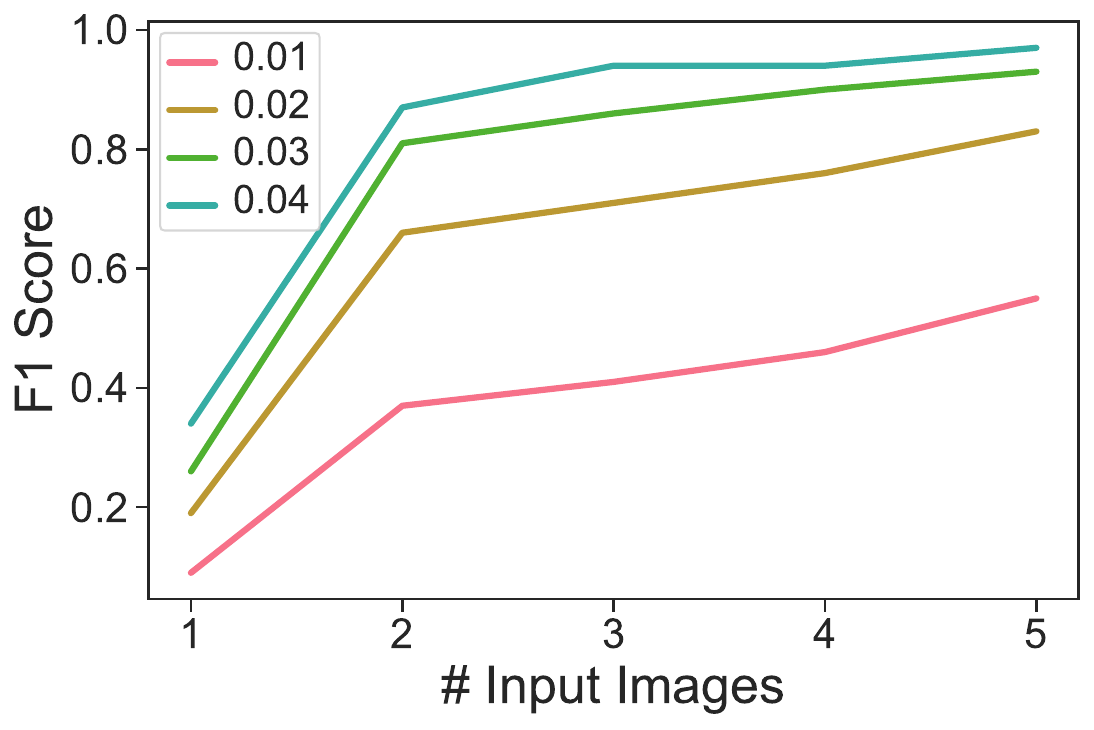}
		\vspace{-0.5em}\captionof{figure}{\textbf{F1-Score for 3D reconstruction per input  set size.} }
		\label{fig:f1}
\end{minipage}
\end{table*}
\paragraph{Losses.}
The loss comprises a data term on the reference images, a 2D diffusion prior term for novel views, and 3D shape regularization terms. 
The data term is a photometric loss between the input images $I_i$ and the rendered image from the corresponding viewpoint $\mathcal{R}_\psi(\pi_i)$:
$$\loss_\text{data} = \mathbb{E}_i \left\| \mathcal{R}^\text{RGB}_\psi(\pi_i) - I^\text{RGB}_i \right\|^2 +  \mathbb{E}_i \left\| \mathcal{R}^\text{Mask}_\psi(\pi_i) - I^\text{Mask}_i \right\|^2$$

The 2D diffusion prior is a score distillation sampling loss \cite{poole2022dreamfusion} adapted to the view-conditioned diffusion model. Intuitively, it guides the renderings from novel viewpoints $\pi$ to be similar to what the diffusion model would predict when conditioned on a randomly selected input view $I_i$:
$$\nabla_\psi \loss_\text{SDS} = \mathbb{E}_{\pi,i,t, \epsilon}\left[\epsilon_{\theta}\left(z_{t}, t, c(\pi/\pi_i), I_i\right) - \epsilon\right] \frac{\partial \mathcal{R}^\text{RGB}_\psi(\pi)}{\partial \psi}$$ 
Here, $z_t$ is the noisy latent encoded from the rendered image $\mathcal{R}_\psi(\pi)$ and $\pi/\pi_i$ is the relative camera pose between $\pi$ and $\pi_i$.
Lastly, we regularize the 3D NeRF being optimized by guiding the normals to be smooth and the rendered masks to be sparse and low-entropy, using a regularization loss $\loss_\text{reg}$ (see the Appendix for more details).

Our final loss is a weighted sum of these losses:
\begin{equation}
    \loss =  \loss_\text{SDS} + \lambda_\text{data}\loss_\text{data} +  \lambda_\text{reg}\loss_\text{reg}
\end{equation}
\begin{figure*}
\centering
    \includegraphics[width=0.90\linewidth]{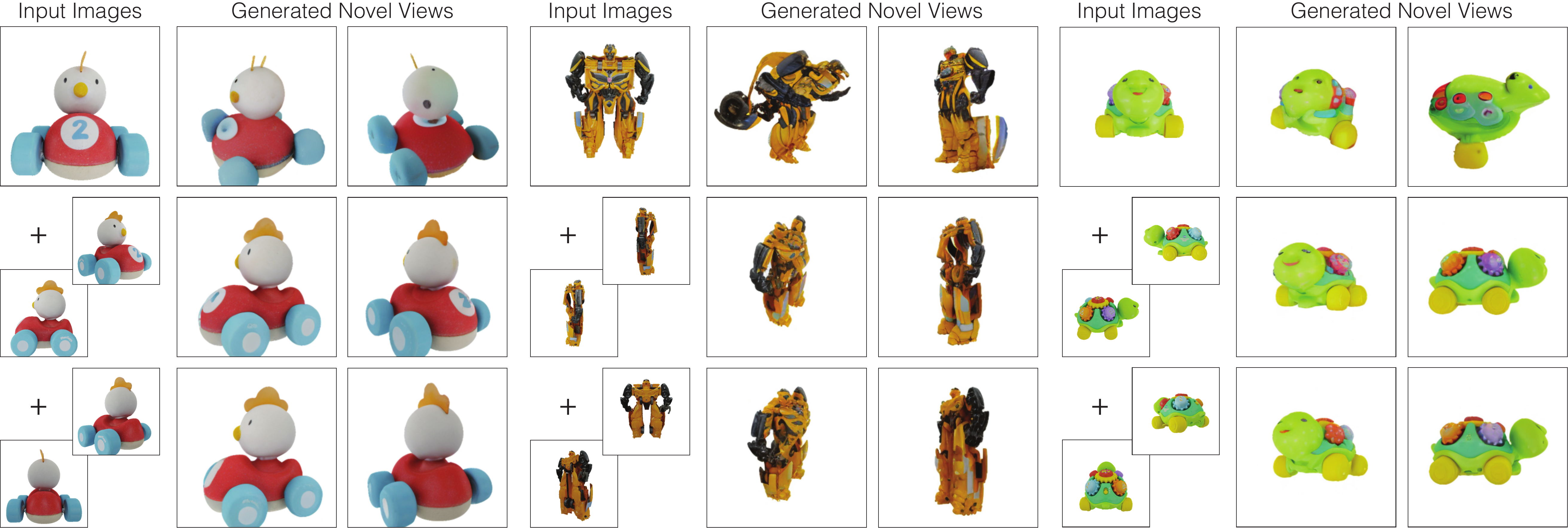}
    \vspace{-0.5em}    \caption{\textbf{\ApproachName~novel view qualitative results.} We present results for 1, 3, and 5 input images. With more input images, \ApproachName~improves fidelity of generated 3D details.}
    \label{fig:nvs}
    \vspace{-1em}
\end{figure*}
\begin{table}
\vspace{-0.7em}
\small
    \centering
    \begin{tabular}{c|ccc}
         \#Images&  LPIPS $\downarrow$ &  PSNR $\uparrow$&  SSIM $\uparrow$\\
         \toprule
         1 (Zero-1-to-3)& 0.23 & 15.2 & 0.79 \\
         \midrule
         2 & 0.15 & 17.6 & 0.83\\
         3 & 0.13 & 18.2 & 0.83\\
        
         4 & 0.10 & 19.4 & 0.85\\
         5 & 0.10 & 19.5 & 0.85\\
         \bottomrule
    \end{tabular}\vspace{-0.5em}
    \caption{\small \textbf{2D benchmark}. We evaluate geometry-free novel-view-synthesis on 20 objects from the GSO dataset. As the number of input images increases the results improve.
    \vspace{-0.5em}}
    \label{tab:more_img_2D}
\end{table}
\vspace{-2.2em}
\section{Experiments}
In this section, we experimentally verify our system \ApproachName. \Cref{sec:exp_impl} provides more details on the implementation, \Cref{sec:exp_views} evaluates how \ApproachName ~effectively ingests a variable number of images , and \Cref{sec:exp_abl} verifies our system design choices and their impact on the quality of 3D reconstruction and 2D novel view synthesis.
\subsection{Implementation details}
\label{sec:exp_impl}
\paragraph{Initial camera poses.} We initialize our camera poses using RelPose++~\cite{lin2023relpose++}. We find that the original model, trained on the Co3D~\cite{co3d} (containing $\sim$19,000 objects), does not generalize well to in-the-wild objects with few views. Therefore, we re-train the model on a larger dataset - Objaverse~\cite{objaverse} ($\sim$800K objects) to obtain more accurate camera poses. In our ablations, we refer to this model as RelPose++*.

\paragraph{View-condition 2D diffusion model.} We use Zero-1-to-3~\cite{liu2023zero} as the view-conditioned diffusion model. This model is pre-trained on Objaverse~\cite{objaverse}. The input camera pose to this model is parameterized by azimuth angle, elevation angle, and scale. We therefore convert the initial camera pose estimations to these parameters, and estimate the absolute elevation of the first camera following~\cite{12345}.

\paragraph{Test-time optimization.} We apply test-time optimization for 1000 steps, with a batch size of 1 and a learning rate of 0.1. For the prior preservation loss, we retrieve 50 nearest neighbors by computing the CLIP-ViT-L similarity to each example in Objaverse training data, and use $\alpha=1.5$. We use a learning rate of 100 for optimizing the camera poses, and further restrict the range of the radius to make optimization more stable. Please see the Appendix for more details.

\paragraph{3D Reconstruction.} We use InstantNGP~\cite{mueller2022instant} to represent our NeRF and train it for 3000 steps with the losses described in \Cref{approach:sds}, $\lambda_\text{data}=500$ and 
and learning rate of 0.01. Every minibatch renders one image for the data loss and one for the SDS loss. For the novel views in the SDS loss, we randomly sample azimuth from $\mathcal{U}(0,360)$ and elevation from $\mathcal{U}(-60,60)$.

\subsection{More Sight, More Insight}
\label{sec:exp_views}
\ApproachName~adapts diffusion-based SoTA single-image reconstruction to incorporate information from multiple views. Here, we demonstrate its effectiveness as the number of views increases, \ie more sight gives more insight into the shape and appearance of the object. We evaluate the output 3D shape and appearance qualitatively and quantitatively. 

\paragraph{Datasets.} We benchmark quantitatively on Google's Scanned Objects (GSO)~\cite{gso}. We randomly selected 20 objects and rendered $k$ views of each object as inputs to our method, $k \in \{1,2,3,4,5\}$. We set the change in the azimuth and the elevation between the sampled random camera poses to be at least 30 degrees. Additionally, we provide qualitative results for real objects that we captured for $k \in \{1,3,5\}$.

\paragraph{Evaluation Metrics.}
To evaluate the appearance of the 3D reconstructions and the synthesized novel views, we use PSNR, SSIM, and LPIPS~\cite{lpips}. For evaluating the geometry of the 3D reconstructions we extract a mesh from the optimized NeRF and benchmark Chamfer Distance, F1 score at different thresholds, and VolumeIoU. For evaluating cameras, we compute the error in relative rotations (in degrees) between all pairs of input cameras, following \cite{lin2023relpose++}.

\paragraph{3D reconstruction improves with more images.} We evaluate the 3D reconstruction quality of \ApproachName. When only one view is provided, no test-time adaptation is performed as we do not have labels to fine-tune the diffusion model.  As shown in \Cref{tab:more_img_3D} and \Cref{fig:f1}, both the geometry and the appearance improve with more input images. The largest improvement is between one view and two views, where test-time optimization is added to the process. The improvements diminish with more views and saturate with 4 input views.  \Cref{fig:real} shows the same trend for 3D reconstruction from real images. When the number of input images is raised from  3 to 5, more fine-grained details are reconstructed correctly (the appearance of the bunny's eyes, the geometry of Snoopy's nose, and the eagle's wings).

\paragraph{2D NVS improves with more images.} We evaluate our method for novel view synthesis. When $k = 1$, our method is similar to sampling from Zero-1-to-3~\cite{liu2023zero}. 
\Cref{tab:more_img_2D} presents the average performance on GSO.
Similarly to the 3D reconstruction, the accuracy of the generated novel views improves with more input images. The relative improvement by adding additional images is larger than for 3D reconstruction, as the instance-specific diffusion model is used directly to sample novel views, instead of via distillation into InstantNGP (that incorporates additional priors in the form of regularization terms). Moreover, the improvements do not saturate with 5 images. 
Qualitative results are presented in \Cref{fig:nvs}. As shown, with more images provided, the diffusion model hallucinates fewer incorrect details (e.g. the sign on the back of the toy car disappears), and generates fine details more accurately (e.g. the parts of the robot).

\paragraph{Camera poses improve with more images.} We evaluate the accuracy of estimated relative camera poses. As shown in \Cref{fig:rot}, the relative camera poses from \ApproachName~improve significantly with more views.

\begin{table*}[h!]
\begin{minipage}{0.65\linewidth}
\small
    \centering
    \begin{tabular}{l|@{\hskip4pt}c@{\hskip4pt}@{\hskip4pt}c@{\hskip4pt}@{\hskip4pt}c@{\hskip4pt}|@{\hskip4pt}c@{\hskip4pt}@{\hskip4pt}c@{\hskip4pt}@{\hskip4pt}c@{\hskip4pt}}
            & LPIPS $\downarrow$& PSNR $\uparrow$ &  SSIM $\uparrow$&  CD $\downarrow$&  F1@0.04 $\uparrow$ &VolumeIoU $\uparrow$\\
         \toprule
            RelPose++& 0.22 & 13.9 & 0.82 & 0.229 & 0.44& 0.41\\
         \midrule
            \ApproachName~w/o adapt.& 0.17 & 17.1 & 0.85 & 0.029 & 0.87& 0.60\\
            \ApproachName~w/o $L_\text{SDS}$ & 0.51 & 9.8 & 0.57 & 0.281 & 0.33& 0.22\\
            \ApproachName~w/o $L_\text{data}$ & 0.20 & 15.3 & 0.84 & 0.030 & 0.76 & 0.52 \\
             \ApproachName~& \textbf{0.16} & \textbf{18.1} & \textbf{0.86} &  \textbf{0.015} & \textbf{0.94} & \textbf{0.63} \\
    \bottomrule    
    \end{tabular}
    \vspace{-0.5em}
    \caption{\small \textbf{\ApproachName~ablations for 3D reconstruction.} We ablate the SDS loss, photometric loss, and test-time adaptation. We evaluate on 13 objects as InstantNGP failed to converge when camera poses were initialized with RP++ for 7 objects.}
    \label{tab:abl_3D}\end{minipage}\hfill
\begin{minipage}{0.30\linewidth}
\centering
		\includegraphics[width=\textwidth]{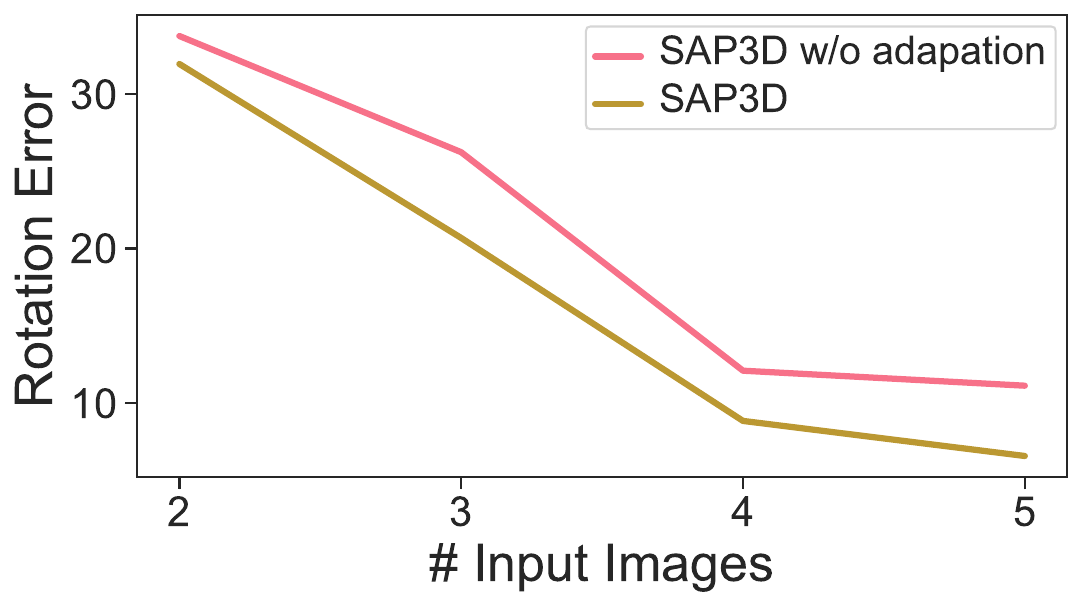}
		\captionof{figure}{\textbf{Camera Pose Evaluation.} The rotation error (in degrees) with and without test-time adaptation for varying numbers of input images.}
		\label{fig:rot}
\end{minipage}

\vspace{-0.6em}
\end{table*}

\begin{figure*}
    \centering
    \begin{minipage}{0.58\textwidth}
        \centering
\includegraphics[width=\textwidth]{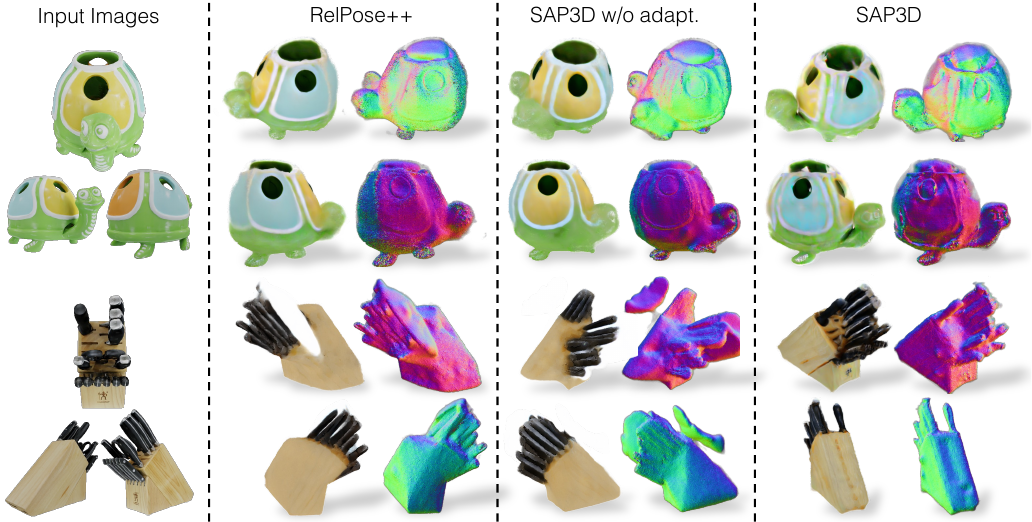} 
        \vspace{-0.5em}\caption{\textbf{Qualitative ablation results for 3D reconstruction.} Training RelPose++ on large-scale data and adapting the view-conditioned diffusion model at test-time improve results.}
        \label{fig:abl_3D}
    \end{minipage}\hfill
    \begin{minipage}{0.40\textwidth}
        \centering
        \includegraphics[width=0.9\textwidth]{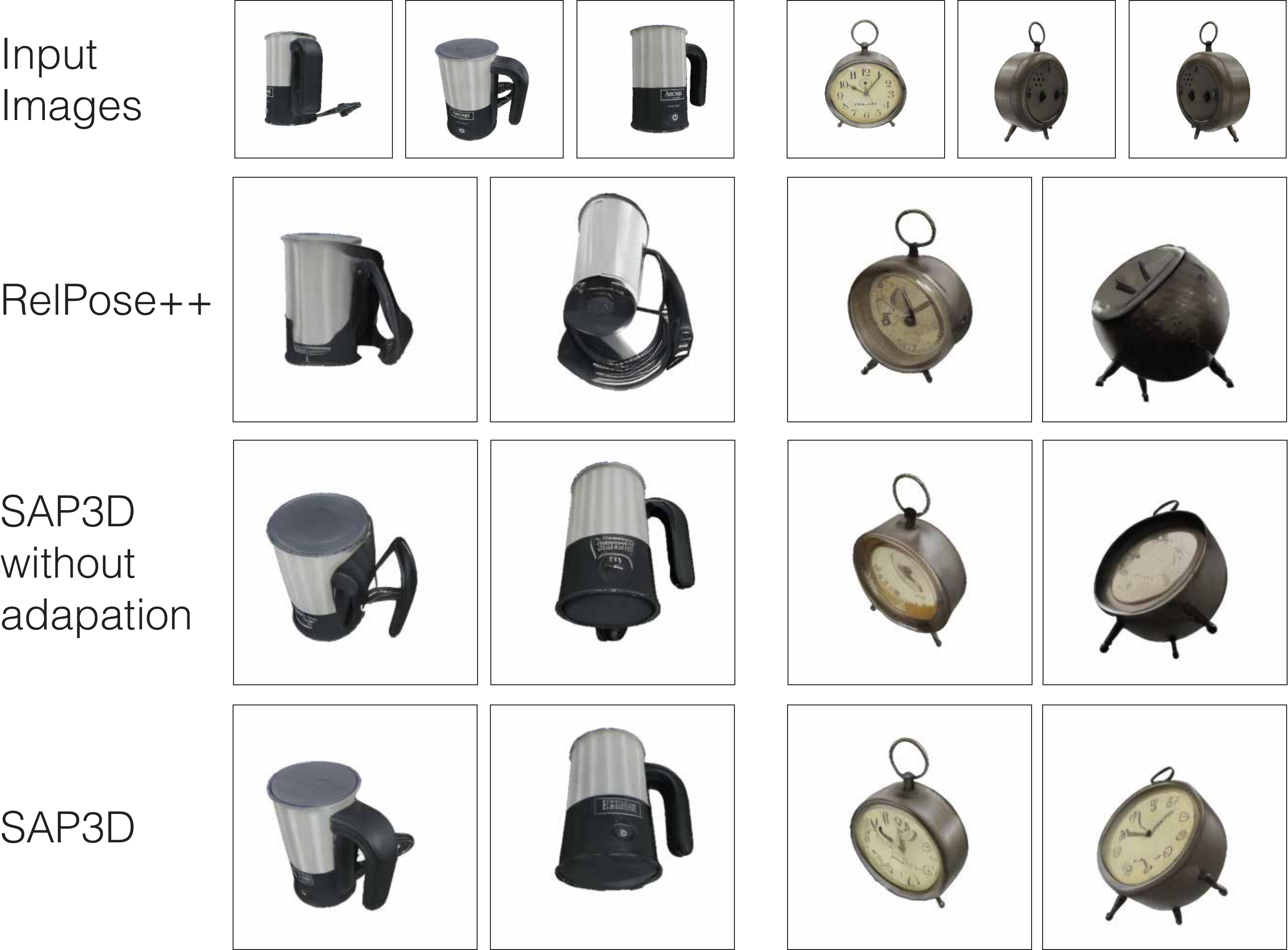} 
        \caption{\textbf{Qualitative ablation results for NVS.}}
        \label{fig:abl_2D}
    \end{minipage}
\end{figure*}

\subsection{System Verification}
\label{sec:exp_abl}
In this section, we ablate different components in our pipeline to verify their effects on the downstream 3D reconstruction (\Cref{tab:abl_3D}) and novel view synthesis performance (\Cref{tab:abl_nvs}). Additionally, we ablate the effect of scaled RelPose++ and the fine-tuning stage on camera poses (\Cref{fig:rot}). 

All of our ablations are done on 20 randomly chosen objects from GSO, and use 3 input images. We incorporate all the 3 images by using stochastic conditioning when sampling from the ablated models.

\paragraph{Initial camera poses ablations.} We compare the downstream 3D reconstruction and novel view synthesis performance with different camera pose initializations. We compare the provided RelPose++ model~\cite{lin2023relpose++} to our version, trained on Objaverse (RelPose++*), before introducing the adaptation stage to the model.  As shown quantitatively in \Cref{tab:abl_3D} (first two rows) and qualitatively in \Cref{fig:abl_3D}, both the geometry and appearance of the downstream 3D reconstruction improve when RelPose++ is replaced with RelPose++*. Similar results hold for NVS downstream evaluation, as shown in \Cref{tab:abl_nvs}. We conclude that large-scale pre-training of the camera pose estimator results in better initialization of the camera poses that improves downstream performance.

\paragraph{Test-time adaptation ablation.} We compare our model with and without test-time training (appears in the tables as ``\ApproachName~w/o adaptation''). Note that without test-time optimization, our system uses the original Zero-1-to-3 model as a prior. As shown in \Cref{tab:abl_nvs}, test-time optimization improves the novel view synthesis in all metrics. As shown in \Cref{tab:abl_3D}, test-time optimization improves 3D reconstruction as well, although the improvement in appearance quality is smaller compared to NVS. Nevertheless, there is a significant improvement in geometry metrics. 
Finally, the test-time adaptation stage results in more accurate camera rotation predictions, as shown in \Cref{fig:rot}. 
\paragraph{3D reconstruction losses ablations.} We ablate the contributions of the data and SDS losses to the 3D reconstruction quality. Without SDS, the 3D model optimization reduces to InstantNGP, and the 2D priors are not used for the reconstruction. As shown in \Cref{tab:abl_3D}, the reconstruction quality is the lowest in this case. The reconstruction quality after applying SDS and MSE is higher than applying each of the losses separately. This suggests that the reconstruction can benefit from applying 2D priors as well as a photometric loss that relies on the predicted relative camera poses of~\ApproachName. 

\section{Discussion and Limitations}

We presented a system that enables 3D reconstruction and generation of novel views from an arbitrary number of images, and improves with more images. We discuss here our two main limitations and conclude with future work. 
\begin{table}[h!]
\small
    \centering
    \begin{tabular}{l|ccc}
             &  LPIPS $\downarrow$&  PSNR $\uparrow$&  SSIM $\uparrow$\\ 
         \toprule
            RelPose++ & 0.30 & 13.3 & 0.77 \\
            \ApproachName~w/o adaptation & 0.18 & 16.3 & 0.78 \\
            \ApproachName &  \textbf{0.13} & \textbf{18.2} & \textbf{0.83} \\
            \bottomrule
    \end{tabular}
    \vspace{-0.5em}
    \caption{\small \textbf{\ApproachName~ablations for novel view synthesis}.} 
    \label{tab:abl_nvs}
\end{table}
\paragraph{Camera pose parametrization.} Our system inherits the camera pose parameterization of the pre-trained diffusion model, restricted to 3 degrees of freedom in the case of \cite{liu2023zero}. We believe that replacing the diffusion model with a generative model that is conditioned on a parametrization with more degrees of freedom will result in more control in novel view synthesis.  

\paragraph{Optimization-based system.} When the number of input images is larger than 1, our system requires an optimization stage of a large-scale diffusion model and therefore can not be applied in real-time. Specifically, the diffusion fine-tuning stage takes around 15 minutes per object on one A100 GPU. 

\paragraph{Future work.} While our current system contains a few independent components, we believe that an end-to-end approach can lead to better performance. Additionally, we hypothesize that model performance can improve at test-time with more inputs, without the need for fine-tuning, by directly training the model to use input examples in context, as done in large language models. We plan to explore this in future work.
{
    \small
    \bibliographystyle{ieeenat_fullname}
    \bibliography{main}
}

\clearpage

\appendix
\section{Appendix}

\subsection{Additional Implementation Details}
\paragraph{Camera pose fine-tuning.} At test time, the camera poses are fine-tuned jointly with the diffusion model, by simply back-propagating gradients to camera parameters, represented as azimuth, elevation, and radius. We update them using a much higher learning rate of $100$, which is $1000\times$ the learning rate of the diffusion model. Additionally, we constrain the camera elevation and scale to make optimization more stable: Elevation is projected to $[0,\pi]$ every iteration, and the radius is mapped to a range of $[1.5, 2.2]$ via a SoftMax.


\paragraph{3D reconstruction.}
We regularize our 3D reconstruction using a number of loss terms. To get smooth surfaces, we regularize surface normals $\hat{n}$ to be smooth. $\loss_{\hat{n},1}$ regularizes normals at sampled 3D points $X$ to be smooth to small perturbations $\delta$, while $\loss_{\hat{n},2}$ regularizes rendered normals $\mathcal{R}^{\hat{n}}_\psi(\pi)$ from random camera viewpoints $\pi$ to be smooth. 
$$ \loss_{\hat{n},1} = \mathbb{E}_{X,\delta\in\mathcal{N}(0,1)} \left\| \hat{n}(X) - \hat{n}(X+\delta) \right\|^2 $$
$$ \loss_{\hat{n},2} = \mathbb{E}_\pi \left\| \Delta \mathcal{R}^{\hat{n}}_\psi(\pi) \right\|^2 \nonumber $$

Additionally, we regularize the density field to form opaque surfaces without floating artifacts. $\loss_\text{Sparse}$ nudges rendered masks $\mathcal{R}^\text{Mask}_\psi(\pi)$ to be sparse with an L1-regularization loss to prevent floaters, while $\loss_\text{Opaque}$ minimizes their entropy to make the closer to 0/1, from random cameras $\pi$.
$$ \loss_\text{Sparse} = \mathbb{E}_\pi \left\| \mathcal{R}^\text{Mask}_\psi(\pi) \right\|_1 $$
$$ \loss_\text{Opaque} = \mathbb{E}_\pi [ H(\mathcal{R}^\text{Mask}_\psi(\pi)) ] $$

Total regularization is a weighted sum of the normal and mask loss terms with $\lambda_{\hat{n},1} = 0.1$, $\lambda_{\hat{n},2} = 0.1$, $\lambda_\text{Sparse} = 1$, and $\lambda_\text{Opaque} = 1$:

$$ \loss_\text{Reg} = \lambda_{\hat{n},1} \loss_{\hat{n},1} + \lambda_{\hat{n},2} \loss_{\hat{n},2} + \lambda_\text{Sparse} \loss_\text{Sparse} + \lambda_\text{Opaque} \loss_\text{Opaque} $$

\subsection{Additional Novel View Synthesis Ablations}

We provide an additional ablation study of the components of our system, evaluated on novel view synthesis. \Cref{tab:ablations_all} presents all of the ablated components.

\paragraph{3D preservation loss.} We compare different versions of regularization losses, applied during the fine-tuning stage of the view-conditioned diffusion model. Specifically, we evaluate three types: no regularization ($\texttt{b2}$), regularization by sampling random pairs from the pre-training set and incorporating them during the fine-tuning ($\texttt{b2}$), and regularization by incorporating nearest-neighbors according to CLIP similarity score (3D reservation loss, as described in \Cref{approach:ttt}). 

As presented in \Cref{tab:ablations_all}, incorporating random images from the training set during the fine-tuning results in comparable performance to not applying regularization. Differently from these two options, using CLIP as a metric for retrieving nearest neighbors from the training data results in an improvement in all metrics. 

\paragraph{Camera initialization and refinement.} Similarly to the ablations presented in \Cref{sec:exp_abl}, we compare the downstream performance with different camera pose initialization - RelPose++ (\texttt{d2}) and RelPose++* (\ApproachName). Differently from \Cref{sec:exp_abl}, we apply the fine-tuning stage in both cases. As shown in \Cref{tab:ablations_all}, scaling the training data of RelPose++ results in significant improvements in novel view synthesis - 4.3dB increase in PSNR. Additionally, the results of initialization from RelPose++* are comparable to initialization with ground-truth camera poses (\texttt{d3}). 

We evaluate the effect of \textit{not} fine-tuning the camera poses together with the view-conditioned diffusion model (\texttt{d1}). The camera-pose fine-tuning that is done in \ApproachName~ results in 0.4dB improvement in PSNR. Therefore, fine-tuning the camera pose is beneficial for downstream novel view synthesis. 

\paragraph{Sampling conditioning.} During the novel view generation process, we use stochastic conditioning - each sampling step from the diffusion model is conditioned on a randomly sampled image from the input images. We compare this conditioning strategy to two different conditioning strategies - using one random image for all the diffusion sampling steps ($\texttt{c2})$, and conditioning the diffusion process on the closest input image, computed according to the camera pose ($\texttt{c1}$). As shown in \Cref{tab:ablations_all}, conditioning on the nearest image results in better performance than conditioning on one random image, and stochastic sampling (\ApproachName) is better than both. 

\subsection{Additional Qualitative Results}
\begin{figure*}[h]
    \vspace{-1em}
  \centering
  \includegraphics[width=0.99\textwidth]{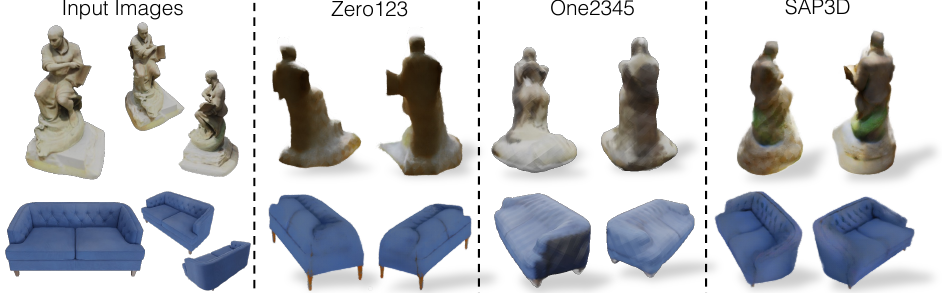}
   \caption{3D reconstruction comparisons on Tanks and Temples and ABO respectively between \ApproachName, Zero123~\cite{liu2023zero} and One2345~\cite{liu2023one2345}.}
   \label{fig:rebuttal-qual-abo-t2}
  \vspace{-1em}
\end{figure*}

In Figure \ref{fig:rebuttal-qual-abo-t2}, We show two sets of qualitative reconstruction comparisons: one on the ABO dataset \cite{collins2022abo} and another on the Tanks and Temples dataset \cite{Knapitsch2017TanksAT}. These comparisons evaluate our proposed method SAP3D against Zero123 \cite{liu2023zero} and One2345 \cite{liu2023one2345}.

We provide additional qualitative results for novel view synthesis and 3D reconstruction, as well as additional qualitative comparisons for our ablation study in \url{https://sap3d.github.io/supp.html}. 

\begin{table*}[h!]
\centering
\begin{tabular}{c|c|c|c|cc||ccc}
\toprule
            & Model     & \multirow{2}{*}{Regularization}& Sampling & \multicolumn{2}{c||}{Cameras}    & \multicolumn{3}{c}{Novel View Quality} \\
            & Finetuned  &                                 &   Conditioning & Initialization& Refine             & PSNR$\uparrow$       & SSIM$\uparrow$           & LPIPS$\downarrow$       \\
\midrule
            \ApproachName & \checkmark  & CLIP-retrieved& Stochastic                    & RelPose++$^*$& \checkmark            & 17.7        & 0.83          & 0.13           \\
\midrule
\texttt{a}  &             &                           & Stochastic                    & RelPose++$^*$&                       & 16.3        & 0.78          & 0.18           \\
\midrule
\texttt{b1} & \checkmark  & Random                    & Stochastic                    & RelPose++$^*$& \checkmark            & 16.8        & 0.79          & 0.16           \\
\texttt{b2} & \checkmark  & None                      & Stochastic                    & RelPose++$^*$& \checkmark            & 16.4        & 0.79          & 0.17            \\
\midrule
\texttt{c1} & \checkmark  & CLIP-retrieved& Nearest                            & RelPose++$^*$& \checkmark            & 16.8        & 0.72          & 0.16           \\
\texttt{c2} & \checkmark  & CLIP-retrieved& Random single & RelPose++$^*$& \checkmark            & 15.2        & 0.79          & 0.23           \\
\midrule
\texttt{d1} & \checkmark  & CLIP-retrieved& Stochastic                    & RelPose++$^*$&                       & 17.3              & 0.80             & 0.15           \\
\texttt{d2} & \checkmark  & CLIP-retrieved& Stochastic                    & RelPose++& \checkmark            & 13.4              & 0.74             & 0.29           \\
\texttt{d3} & \checkmark  & CLIP-retrieved& Stochastic                    & Ground Truth& \checkmark            & 17.8              & 0.84             & 0.13           \\
\bottomrule
\end{tabular}
  \caption{\textbf{Ablation study on novel view synthesis.} We evaluate the effect of various design choices on novel view synthesis. RelPose++$^*$ denotes our RelPose++ model trained on Objaverse. In the paper, we refer to $\texttt{a}$ as ``\ApproachName~w/o adaptation''.  Please see the text for details. }
  \label{tab:ablations_all}
\end{table*}

\paragraph{Acknowledgments.} \looseness=-1 The authors would like to thank Yutong Bai for the helpful discussions. YG is funded by the Google Fellowship.

\end{document}